\documentclass[a4paper,fleqn]{cas-dc}

\usepackage[round,authoryear]{natbib}
\usepackage{placeins}
\begin{document}
\let\WriteBookmarks\relax
\def\floatpagepagefraction{1}
\def\textpagefraction{.001}

\shorttitle{Generative Diffusion Models for Agricultural AI}

\shortauthors{Da Tan et~al.}

\title [mode = title]{Generative Diffusion Models for Agricultural AI: Plant Image Generation, Indoor-to-Outdoor Translation, and Expert Preference Alignment}                      
\tnotemark[1]


%\tnotetext[1]{This document is the results of the research
%   project funded by the University of Manitoba.}

% \tnotetext[2]{The second title footnote which is a longer text matter
%    to fill through the whole text width and overflow into
%    another line in the footnotes area of the first page.}

\author[1]{Da Tan}[type=author,
                        auid=000,bioid=1,
                        %prefix=Sir,
                        %role=Researcher,
                        orcid=0009000078616786]

\cormark[1]


\ead{tand2@myumanitoba.ca}


%\credit{Conceptualization of this study, Methodology, Software}

\affiliation[1]{organization={University of Manitoba},
    addressline={400 University Centre}, 
    city={Winnipeg},
    % citysep={}, % Uncomment if no comma needed between city and postcode
    postcode={R3T 2N2}, 
    country={Canada}}

\author[2]{Michael Beck}
\ead{m.beck@uwinnipeg.ca}

\author[2]{Christopher P. Bidinosti}
\ead{c.bidinosti@uwinnipeg.ca}

\author[1]{Robert H. Gulden}
\ead{Rob.Gulden@umanitoba.ca}

\author[1]{Christopher J. Henry}[orcid=0000000226241502]
\ead{christopher.henry@umanitoba.ca}
\cormark[1]
\affiliation[2]{organization={University of Winnipeg},
    addressline={599 Portage Ave}, 
    city={Winnipeg},
    % citysep={}, % Uncomment if no comma needed between city and postcode
    postcode={R3B 2G3}, 
    country={Canada}}
%\credit{Data curation, Writing - Original draft preparation}

\cortext[cor1]{Corresponding author}
%\cortext[cor2]{Principal corresponding author}

%\fntext[fn1]{This is the first author footnote. but is common to third author as well.}
% \fntext[fn2]{Another author footnote, this is a very long footnote and
%   it should be a really long footnote. But this footnote is not yet
%   sufficiently long enough to make two lines of footnote text.}

% \nonumnote{This note has no numbers. In this work we demonstrate $a_b$
%   the formation Y\_1 of a new type of polariton on the interface
%   between a cuprous oxide slab and a polystyrene micro-sphere placed

\begin{abstract}
The success of agricultural artificial intelligence (AI) depends heavily on large, diverse, and high-quality plant image datasets, yet collecting such data in real field conditions is costly, labor-intensive, and seasonally constrained. This paper investigates diffusion-based generative modeling to address these challenges through plant image synthesis, indoor-to-outdoor translation, and expert preference–aligned fine-tuning. First, a Stable Diffusion v1.4 (SD-v1.4) model is fine-tuned on captioned indoor and outdoor plant imagery to generate realistic, text-conditioned images of canola and soybean. Evaluation using Inception Score (IS), Fréchet Inception Distance (FID), and downstream phenotype classification shows that synthetic images effectively augment training data and improve accuracy. Second, we bridge the gap between high-resolution indoor datasets and limited outdoor imagery via DreamBooth-based text inversion and image-guided diffusion, generating translated images that enhance weed detection and classification with YOLOv8. Finally, a preference-guided fine-tuning framework trains a reward model on expert scores and uses reward-weighted updates to produce more stable, expert-aligned outputs. Together, these components demonstrate a practical pathway toward data-efficient, generative pipelines for agricultural AI.
\end{abstract}

% \includegraphics{figs/grabs.pdf}

%%%%%%%%%%%%%%%% Research highlights
% \begin{highlights}
% \item Research highlights item 1
% \item Research highlights item 2
% \item Research highlights item 3
% \end{highlights}

% Keywords

\maketitle
%%%%%%%%%%%%%%%%%%%%%%%%%%%%
\section{Introduction}
Deep learning for digital agriculture relies heavily on large, diverse, and well-annotated image datasets~\citep{kamilaris2018deep}. However, creating such datasets in real fields is expensive, time-consuming, and constrained by seasons and the need for expert labeling. Diffusion models (DMs) ~\citep{sohl2015deep, ho2020denoising} offer a new way to alleviate this bottleneck by generating high-quality synthetic images that can theoretically serve as labeled training data rather than merely visual illustrations. In plant science and digital agriculture, we would like not only to generate additional examples of specific crops and phenotypes, but also to translate abundant greenhouse images into realistic outdoor field conditions~\citep{krosney2023inside}, thereby bridging the domain gap that currently limits model generalization from indoor to outdoor. At the same time, photorealism alone is not sufficient. Many ``good-looking'' images fail to satisfy expert criteria for task-domain specificity and/or botanical realism. This motivates a preference-guided fine-tuning stage, where expert feedback is distilled into a reward model that steers the diffusion process toward outputs that are more consistent and structural meaningful. In this work, we bring these three ideas together: text-conditioned synthetic image generation, indoor-to-outdoor translation, and preference-aligned model fine-tuning for agricultural imagery and downstream machine learning tasks.

Early research in agricultural image augmentation relied on classical techniques such as geometric transformations (rotation, flipping, scaling) or color jittering to expand dataset diversity~\citep{pawara2017data}. While these operations provide marginal improvements, they fail to capture realistic variability in illumination, background, and texture. 

The introduction of generative adversarial networks (GANs) marked a breakthrough in realistic image synthesis~\citep{goodfellow2014generative,krosney2023inside}. GAN-based methods have been applied to agricultural tasks such as plant disease augmentation~\citep{han2025plant} and lab-to-field domain translation~\citep{krosney2023inside}. CycleGAN~\citep{zhu2017unpaired} and Pix2Pix~\citep{isola2017image} have enabled unpaired and paired image-to-image translation, respectively, helping bridge visual gaps between synthetic and real domains. However, GANs are notoriously difficult to train and often produce unstable or low-fidelity outputs, particularly when domain data are scarce~\citep{ahmad2025understanding}. Moreover, GANs lack semantic controllability, making it challenging to specify high-level scene attributes such as plant health status or environmental context through text or structured prompts.

Diffusion models have recently emerged as a powerful alternative for image generation, surpassing GANs in controllability. Denoising diffusion probabilistic models (DDPMs)~\citep{ho2020denoising} and latent diffusion models (LDMs)~\citep{rombach2022high} formulate image synthesis as an iterative denoising process, progressively refining random noise into coherent images guided by learned priors. Unlike GANs, diffusion models are stable to train and can incorporate multi-modal conditioning, such as textual prompts through CLIP-based encoders~\citep{radford2021learning}. These advancements enable a flexible text-to-image generation paradigm, where semantic attributes can be explicitly controlled. Such models have shown strong generalization across diverse domains, but adaptation to agriculture remains underexplored due to domain-specific gaps.

Recent reviews summarize both the promise and current gaps in generative augmentation for agriculture. Lu \textit{et al.}~\citeyearpar{lu2022generative} highlight progress in GAN-based augmentation for crop disease detection, weed segmentation, and yield prediction, while also noting instability and domain-generalization issues. Rahman \textit{et al.}~\citep{ur2024generative} emphasize that most agricultural generative studies remain small scale and crop specific. 

To address the data gap, we propose a unified diffusion-based framework comprising three components: \textbf{(1) text-conditioned plant image generation}, \textbf{(2) indoor-to-outdoor image translation}, and \textbf{(3) expert preference-aligned model fine-tuning}.

First, we fine-tune Stable Diffusion v1.4 (SD-v1.4) on captioned indoor and outdoor crop datasets. Image-caption pairs are generated using ChatGPT-4o~\citep{openai_gpt4o_2024}, enabling prompts that reflect agriculturally relevant semantics (\textit{e.g.}, ``canola with yellowing leaves under drought stress''). 

Second, we develop an indoor-to-outdoor translation pipeline. For single-object scenes, we adopt a DreamBooth-based text-conditioning strategy~\citep{ruiz2023dreambooth} to bind the indoor plant structure to a unique identifier token, enabling prompt-driven translation into outdoor scenes. For multi-object scenes, we instead use an image-guided diffusion method~\citep{meng2021sdedit}, which preserves spatial composition of multiple plants in translation.

Third, we introduce a preference-guided fine-tuning stage to improve perceptual stability and realism. A convolutional neural network (CNN)-based reward model is trained on expert-scored generations to predict preference signals, which are used in a reward-weighted supervised fine-tuning objective. This aligns the generative model with expert preferences.

Across multiple downstream experiments, including phenotype classification and weed detection, we demonstrate that synthetic and translated images improve downstream performance. Preference-aligned fine-tuning further enhances coherence and realism for output images. 

The rest of this paper is organized as follows. Section~2 outlines the theoretical preliminaries. Section~3 presents the overall methodology. Sections~4, 5, and 6 describe the three core components of this study: text-conditioned image generation, indoor-to-outdoor image translation, and expert-preference-guided model fine-tuning, respectively. Section~7 provides an overall discussion, and Section~8 concludes the paper.

%%%%%%%%%%%%%%%%%%%%%%
\section{Theoretical Preliminaries}
This section introduces the theoretical preliminaries of this research, notably the variational autoencoder, diffusion and latent diffusion models, and DreamBooth for learning representation of a single object.

\subsection{Variational Autoencoders}
A variational autoencoder (VAE) learns to compress and reconstruct an input image $x$ using an encoder $E(\cdot)$ and a decoder $D(\cdot)$ by maximizing the evidence lower bound (ELBO):
\begin{equation}
    \mathcal{L}_{\text{VAE}} 
    = \mathbb{E}_{q_\phi(z|x)}\left[\log p_\theta(x|z)\right] 
    - D_{\text{KL}}\left(q_\phi(z|x)\,\|\,p(z)\right),
\end{equation}
where $z$ denotes the latent variable, $q_\phi(z|x)$ is the approximate posterior distribution parameterized by encoder parameters $\phi$, $p_\theta(x|z)$ is the likelihood of decoded output parameterized by $\theta$, and $p(z)=\mathcal{N}(0,I)$ is a standard Gaussian prior. 
$D_{\text{KL}}(\cdot\|\cdot)$ denotes the Kullback--Leibler divergence~\citep{kullback1951information}. The first term encourages accurate reconstruction of the input, while the second term regularizes the latent distribution toward the prior.

\subsection{Diffusion Models and Latent Diffusion Models}
Diffusion models~\citep{ho2020denoising} learn to gradually reverse a Markovian noise corruption process applied to an input image. 
During training, a data sample $x_0 \in \mathbb{R}^{H \times W \times C}$ is progressively corrupted through a forward diffusion process:
\begin{equation}
    q(x_t \mid x_{t-1}) 
    = \mathcal{N}\!\left(x_t;\, \sqrt{1-\beta_t}\,x_{t-1},\, \beta_t I\right),
\end{equation}
where $x_t$ denotes the noisy image at diffusion timestep $t \in \{1,\dots,T\}$. $q(x_t \mid x_{t-1})$ is the conditional probability of $x_t$ given $x_{t-1}$. It is a Gaussian distribution and no trained model is involved in this step. $\beta_t \in (0,1)$ is a predefined noise variance schedule, and $I$ is the identity matrix. The reverse diffusion process is parameterized by a neural network with parameters $\theta$ and learns to iteratively remove noise:
\begin{equation}
    p_\theta(x_{t-1} \mid x_t) 
    = \mathcal{N}\!\left(x_{t-1};\, \mu_\theta(x_t, t),\, \Sigma_\theta(x_t, t)\right),
\end{equation}
where $p_\theta(x_{t-1} \mid x_t)$ denotes the learned conditional distribution for a single denoising step, and $\mu_\theta(\cdot)$ and $\Sigma_\theta(\cdot)$ represent the predicted mean and variance of the reverse process, respectively.
In practice, the model is trained using a simplified denoising objective that minimizes the mean squared error between the true noise and the noise predicted by the model:
\begin{equation}
    \mathcal{L}_{\text{DDPM}} 
    = \mathbb{E}_{x_0,\, \epsilon,\, t}
    \left[ \left\| \epsilon - \epsilon_\theta(x_t, t) \right\|^2 \right],
\end{equation}
where $\epsilon \sim \mathcal{N}(0,I)$ is the Gaussian noise added during the forward process, $\epsilon_\theta(x_t,t)$ denotes the noise predicted by the neural network, and the expectation is taken over the data distribution, noise samples, and uniformly sampled timesteps.

The diffusion model’s denoising network is typically implemented as a U-Net~\citep{ronneberger2015u}, a convolutional encoder-decoder with skip connections. The encoder extracts hierarchical features across spatial scales, while the decoder reconstructs the image through learned upsampling. Skip connections preserve spatial information lost during downsampling, which is crucial for detailed texture recovery in images. 

LDM is a combination of the efficiency of VAE and generative power of diffusion models ~\citep{rombach2022high}. Instead of performing diffusion in pixel space, LDMs operate in the VAE’s latent space, significantly reducing computational cost. The reverse process learns to predict the added noise $\epsilon$ conditioned on text embeddings derived from a pretrained CLIP encoder~\citep{radford2021learning}. 

\subsection{DreamBooth}
DreamBooth~\citep{ruiz2023dreambooth} enables personalized adaptation of diffusion models by introducing a rare identifier token (\textit{e.g.} \texttt{sks})---chosen to avoid interference with existing semantic concepts in the pretrained vocabulary---to associate with the visual concept of a specific object. The model is fine-tuned on a small set of augmented views of the same object, optimizing a composite objective that combines subject reconstruction loss $\mathcal{L}_{\text{subj}}$ and class prior preservation loss $\mathcal{L}_{\text{prior}}$:
\begin{equation}
\label{eq:dreambooth}
    \mathcal{L}_{\text{DreamBooth}} = \mathcal{L}_{\text{subj}} + \lambda\mathcal{L}_{\text{prior}},
\end{equation}
where $\mathcal{L}_{\text{subj}}$ ensures that the generated image preserves subject identity, while $\mathcal{L}_{\text{prior}}$ maintains generalization to the broader object class, preventing overfitting. $\lambda$ is a weighting hyperparameter controlling the trade-off between subject fidelity and generalization. 

\begin{figure*}[h]
    \centering
    \includegraphics[width=\linewidth]{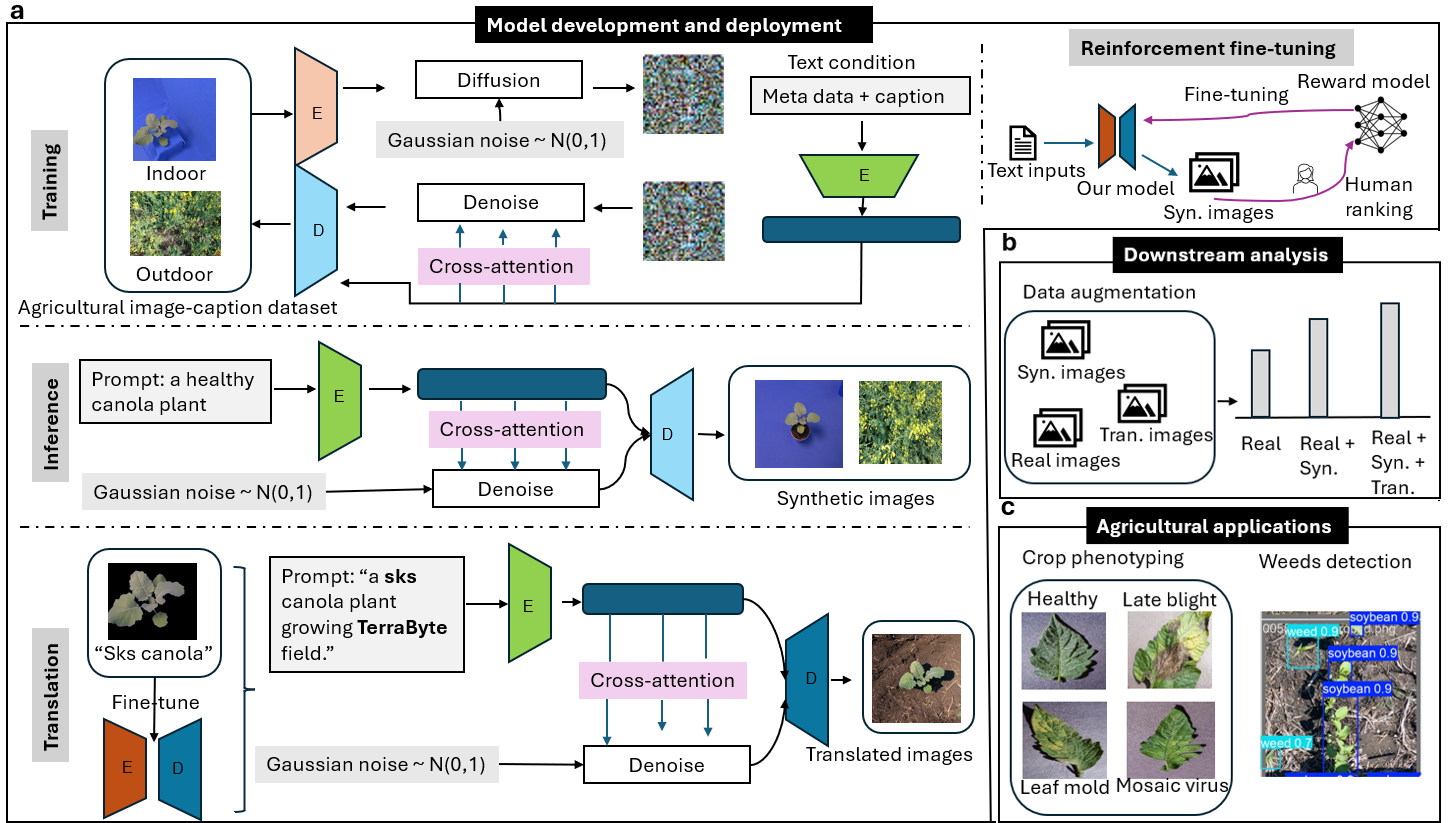}
   \caption{\textbf{Overview of the proposed framework.}  
    (a) Diffusion-based generative workflow encompassing text-conditioned image generation, image translation, and preference-guided fine-tuning.  
    (b) Data augmentation process integrating synthetic and translated images for downstream tasks.  
    (c) Agricultural applications including crop disease phenotyping and weed detection.}
    \label{fig:overall_workflow}
\end{figure*}

\subsection{Overview of Methodology}
The overall workflow of the proposed framework is shown in Figure~\ref{fig:overall_workflow}.  
It unifies three components: text-conditioned image generation, indoor-to-outdoor image translation, and preference-guided fine-tuning, into a single diffusion-based pipeline for agricultural image synthesis and data augmentation. 

Panel (a) highlights the generative process.  
Indoor and outdoor plant images are paired with captions, encoded into the latent space of a LDM, and conditioned on text embeddings via cross-attention. At inference, the model generates synthetic plant images from Gaussian noise based on prompts such as “a healthy canola plant.” For translation, a DreamBooth-style text condition links each indoor plant to a rare identifier token (\texttt{sks}), enabling prompts like “a sks canola plant in outdoor field” to preserve plant morphology while adapting lighting, soil, and background in outdoor scenes. Finally, an expert preference-guided fine-tuning stage uses a reward model trained on expert-scored samples to weight the LDM denoising loss, improving realism and output stability.

Panel (b) illustrates how these synthetic and translated images support downstream machine-learning tasks. By mixing real and generated data, the overall dataset becomes more diverse, improving model robustness in tasks such as phenotype classification and object detection.

Panel (c) shows representative agricultural applications.  
Generated diseased-leaf images enrich phenotype diversity for crop classification, while translated soybean–weed scenes improve YOLO-based ~\citep{yolov8} weed detection and classification models.

To summarize, this study will present a unified diffusion-based framework (Figure~\ref{fig:overall_workflow}) for agricultural data generation that consists of three components:  
(1) text-conditioned plant image generation,  
(2) DreamBooth-based image translation, and  
(3) preference-guided model refinement.  
Each component builds upon a fine-tuned SD-v1.4 backbone~\citep{rombach2022high} adapted to the agricultural domain. The following sections will discuss these components separately.

%%%%%%%%%%%%%%%% 
\section{Text-Conditioned Image Generation}
In this section, we describe the methodology and results of fine-tuning SD-v1.4 for plant image generation. Our focus is on adapting the model to the agricultural domain and systematically evaluating the utility of the generated images in downstream classification tasks. This provides insight into the potential of generative models as data augmentation tools for advancing agricultural machine learning.

\subsection{Datasets and Preprocessing}
Indoor plant images were collected using the EAGL-I system developed by Beck \textit{et al.}~\citep{beck2020embedded}, which captures high-resolution images of plants in controlled laboratory environments (\ref{fig:indoor_examples_canola}). The indoor dataset features standardized lighting conditions and a uniform blue background. Each image is accompanied by metadata describing plant species, developmental stage, and imaging parameters. In this work, we focus on canola (\textit{Brassica napus} L.) and soybean (\textit{Glycine max} [L.] Merr.), two dominant broadleaf crops in the region, as representative crop species for generative modeling and downstream evaluation.

Outdoor field images were obtained from the TerraByte research group~\citep{terabyte2024}, collected at EMILI’s Innovation Farm in Grosse Isle, Manitoba, Canada. These images reflect realistic agricultural field conditions, including variable illumination, background clutter, and occlusions. Data acquisition was performed using GoPro cameras mounted on a tracker boom, resulting in diverse viewpoints and developmental stages. Due to the uncontrolled environment, some images exhibit artifacts such as motion blur, harsh shadows, background machinery, or empty field scenes. Examples of low-quality outdoor images can be found in \ref{fig:low_qua_outdoor_examples}.

To enable text-conditioned training of the diffusion model, image–caption pairs were generated for both indoor and outdoor datasets. Captions were produced using the multimodal large language model ChatGPT-4o~\citep{openai_gpt4o_2024}, which was selected after comparison with BLIP-2~\citep{li2023blip2} and LLaVA~\citep{liu2023llava} due to its superior accuracy in describing fine-grained plant attributes. Prompt engineering was applied to emphasize biologically relevant information such as species, developmental stage, visible stress symptoms, and environmental context (indoor versus field). Caption quality was verified through manual inspection on a representative subset of images. Examples of prompt template and generated image-caption pairs are shown in \ref{fig:prompt}.
 
The captioned dataset was then resized to a fixed resolution of $512 \times 512$ pixels and split into training and test sets using a 90/10 ratio. The final training set comprised approximately 8{,}000 canola images and served as the foundation for both text-conditioned image generation and subsequent indoor-to-outdoor translation experiments.

\subsection{Text-Conditioned Diffusion Model Fine-Tuning}
 We first fine-tuned a pretrained SD-v1.4 model on domain-specific indoor and outdoor canola datasets to enable text-conditioned plant image generation. The pipeline, illustrated in Figure~\ref{fig:overall_workflow}(a), integrates image–caption pairs, prompt embeddings, and latent diffusion denoising.  
Each caption describes both morphological and environmental attributes of plants, providing textual conditioning during generation. The SD-v1.4 architecture was initialized with pretrained weights from the LAION-5B dataset~\citep{schuhmann2022laion}. The model architecture follows the latent diffusion paradigm, consisting of (1) a VAE encoder–decoder that compresses and reconstructs plant images in a 4-channel latent space, (2) a CLIP text encoder~\citep{radford2021learning} that embeds descriptive prompts, and (3) a U-Net denoiser~\citep{ronneberger2015unet} that iteratively refines Gaussian noise into realistic images conditioned on text.

To adapt the diffusion model to the agricultural domain, we fine-tuned only the U-Net denoising backbone of the latent diffusion model, while keeping both the VAE and the text encoder frozen. This strategy significantly reduces computational cost and mitigates catastrophic forgetting of the general visual representations learned in the pretrained model. Fine-tuning was performed using paired image–caption samples from the captioned canola dataset, optimizing the standard DDPM objective in which the U-Net learns to predict the Gaussian noise added to latent representations. Text conditioning was provided through CLIP-based embeddings, enabling the model to align visual generation with biologically meaningful prompt descriptions.

During inference, image synthesis begins by sampling a latent variable from a standard Gaussian prior and iteratively applying the learned reverse diffusion process to obtain a denoised latent representation. The final latents are then decoded back into pixel space using the frozen VAE decoder. 

Fine-tuning was conducted on three NVIDIA A100 GPUs at the University of Manitoba for a total of 3{,}000 training steps. A batch size of 20 was used with gradient accumulation over four steps to achieve a larger effective batch size within GPU memory constraints. The learning rate was set to $1 \times 10^{-5}$, and all images were resized to a resolution of $512 \times 512$ pixels. Early stopping was guided by qualitative inspection: at regular intervals, the model generated samples from fixed validation prompts, which were visually assessed for realism, diversity, and consistency with the conditioning text.

\subsection{Evaluation Metrics and Baseline Comparison}
We evaluated the realism and diversity of generated images using two standard metrics in generative modeling: Fréchet Inception Distance (FID) \citep{heusel2017gans} and Inception Score (IS) \citep{salimans2016improved}. 
FID quantifies the distance between feature distributions of generated and real images extracted from a pretrained Inception network, with lower values indicating closer alignment and higher visual fidelity. 
IS measures both sample quality and diversity by computing divergence between conditional and marginal label distributions predicted by the pretrained Inception model, where higher scores indicate more diverse and semantically coherent generations. 

For benchmarking, we compared our fine-tuned SD-v1.4 model against three representative generative models. Imagen \citep{saharia2022photorealistic} is a large-scale diffusion-based model known for photorealistic text-to-image synthesis but is trained in raw pixel space. 
DALL$\cdot$E \citep{ramesh2021zero} employs a transformer-based architecture with discrete latent representations. 
GigaGAN \citep{kang2023gigagan} is a GAN-based model optimized for high-resolution image generation. 

\subsection{Image Generation Results}
The generated images were evaluated by both visual inspection and quatitative metrics.

\subsubsection{Qualitative Results}

Figure~\ref{fig:imagegen_results}(a) presents examples of generated canola plant images under both indoor and outdoor environmental settings. The indoor samples exhibit high visual fidelity and diversity, capturing fine-grained details such as leaf texture, lighting variations and developmental stages. Meanwhile, the outdoor samples demonstrate strong realism and controllability, with image semantics modulated through text prompts. The generation process can capture distinct developmental stages guided by text prompt, such as those corresponding to prompt key words such as ``June 21'' and ``July 22'' (Figure~\ref{fig:imagegen_results}(a)), illustrating the model’s ability to synthesize temporally consistent plant development under varying field conditions.

\subsubsection{Quantitative Evaluation}
We evaluated the generated images using the FID and IS, as well as their impact on a downstream plant disease classification task which will be mentioned in later sections. Table~\ref{tab:is_fid} summarizes the metrics, and the bar plots (Figure~\ref{fig:imagegen_results}(b)) visualize comparative performance. 

For the IS score, our model achieves the highest score of 3.29 on indoor images, outperforming Imagen (3.04), DALL$\cdot$E (2.89), and GigaGAN (2.44). On the outdoor dataset, Imagen slightly outperforms our model (2.72 vs. 2.60), while DALL$\cdot$E (2.21) and GigaGAN (2.06) achieve lower scores. The overall trend indicates that indoor images yield higher IS values than outdoor images, likely due to the reduced complexity and controlled conditions of laboratory settings.  

For FID, our model consistently achieves the best results in both domains. On the indoor dataset, we obtain a score of 83.4, compared to GigaGAN (98.4), DALL$\cdot$E (103.9) and Imagen (118.2). On the outdoor dataset, our approach again performs best with 127.6, followed by DALL$\cdot$E (129.2), GigaGAN (133.1), and Imagen (149.1).  

Although all models show higher FID values for outdoor images—indicating increased difficulty in modeling field variability—our fine-tuned diffusion model demonstrates clear advantages in both fidelity and diversity. 

\begin{table}[h]
\centering
\caption{Comparison of IS and FID for generated canola plant images.}
\label{tab:is_fid}
\begin{tabular}{l|cc|cc}
\hline
\multirow{2}{*}{Model} & \multicolumn{2}{c|}{IS $\uparrow$} & \multicolumn{2}{c}{FID $\downarrow$} \\
\cline{2-5}
 & Indoor & Outdoor & Indoor & Outdoor \\
\hline
Imagen   & 3.04 & \textbf{2.72} & 118.20 & 149.10 \\
GigaGAN  & 2.44 & 2.06 & 98.44 & 133.06 \\
DALL$\cdot$E    & 2.89 & 2.21 & 103.89 & 129.21 \\
Ours     & \textbf{3.29} & 2.60 & \textbf{83.40} & \textbf{127.60} \\
\hline
\end{tabular}
\end{table}

\subsection{Downstream Machine Learning Evaluation}
To further validate the effectiveness of our image generation model, we designed a downstream machine learning task in which synthetic images are used for data augmentation. The objective was to assess whether the inclusion of generated images improves the performance of phenotype classification models compared to training solely on real data. 

\subsubsection{Dataset and Experiment Setup}
We conducted phenotype classification experiments on two widely used plant disease benchmarks: PlantVillage \citep{mohanty2016using} and CropDisease \citep{mensah2023ccmt}. 
PlantVillage contains leaf images of tomato, potato, and bell pepper across multiple healthy and diseased phenotypes, while CropDisease includes maize, cashew, and cassava with a diverse set of disease and pest categories. For all species, datasets were split into training and test sets using a 70/30 ratio, and classification models were trained independently for each crop to assess the impact of synthetic augmentation across different phenotypes.

For each crop species, a pretrained SD-v1.4 model was fine-tuned using plant image–caption pairs derived from the respective training sets. Image captions were generated using LLaVA-1.5 \citep{liu2023llava}, which provided sufficient descriptive accuracy for the relatively simple leaf imagery while remaining computationally efficient. Following fine-tuning, synthetic images were generated from captions corresponding to the test set. To avoid data leakage, the test set was further split into 60\%/40\%, where synthetic images were generated only from the 60\% portion, and all evaluations were performed exclusively on the remaining 40\% real images. Details of training data are shown in Appendix \ref{tab:tomato-potato-bellpepper} and \ref{tab:maize-cashew-cassava}.

Phenotype classification was performed using a lightweight custom CNN architecture shared across all crops. The network consists of four convolutional blocks with batch normalization and ReLU activations, followed by progressive downsampling and an adaptive average pooling layer. This design balances representational capacity and computational efficiency, making it suitable for small-to-medium agricultural datasets. A fully connected classifier with dropout regularization maps extracted features to phenotype labels. Details of the CNN architecture is shown in \ref{tab:cnn-architecture-compact}.

To quantify the effect of synthetic data augmentation, classification models were trained under varying synthetic-to-real data ratios of 0\%, 50\%, 100\%, 200\%, 300\%, and 400\%. 
At 0\%, only real images were used, while higher ratios progressively added synthetic samples while keeping lower-ratio samples fixed. All synthetic images were drawn from a common generation pool to ensure fair comparison across augmentation levels.

\subsubsection{Results and Analysis}
\begin{figure*}[h] %[t]
    \centering
    \includegraphics[width=\linewidth]{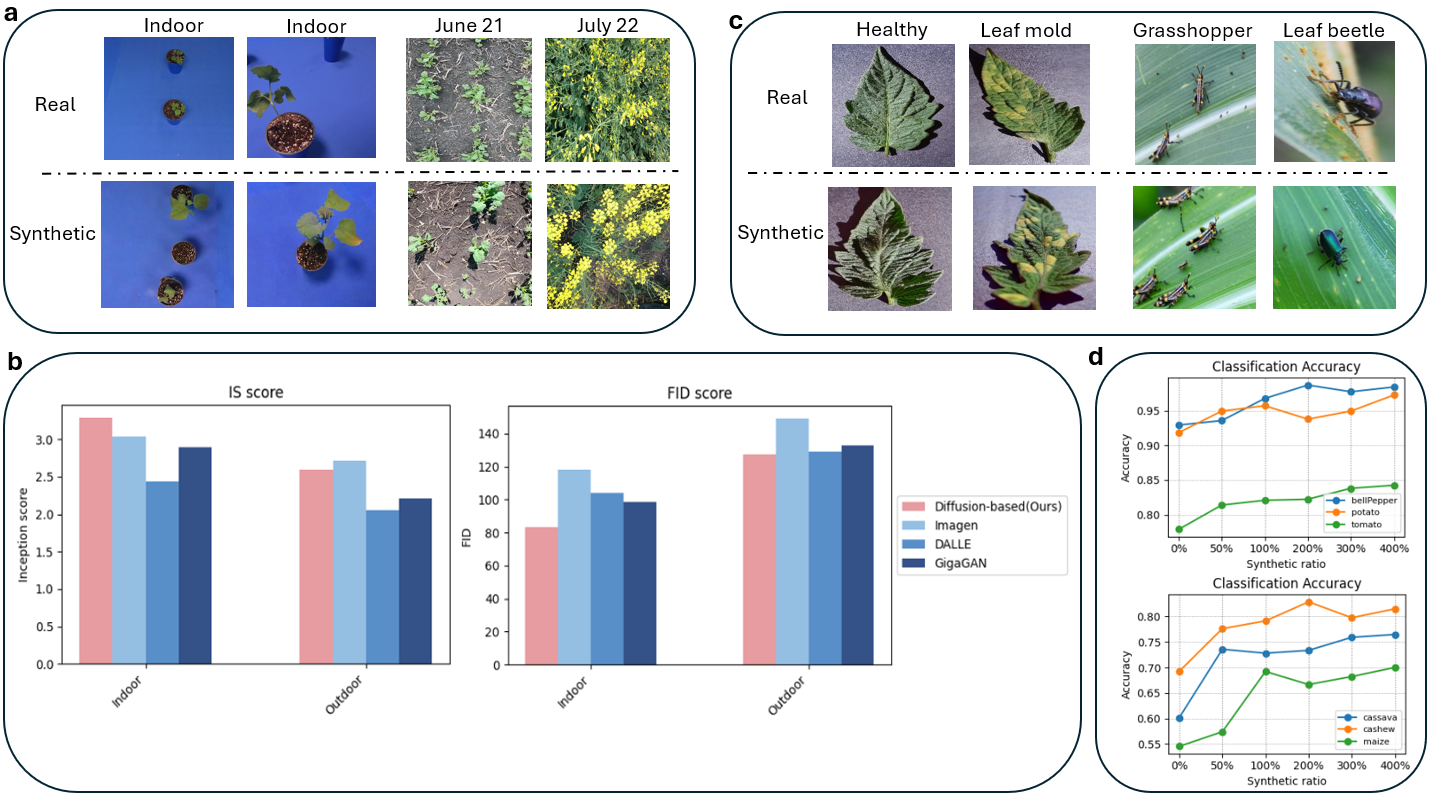}
    \caption{\textbf{Text-conditioned image generation results.}  
    (a) Comparison between generated and real images for both indoor and outdoor canola plants.  
    (b) Inception Score (IS) and Fréchet Inception Distance (FID) comparison with three baselines (Imagen, GigaGAN, DALL·E) for indoor and outdoor datasets.  
    (c) Examples of generated vs.\ real images for downstream machine learning datasets: tomato (PlantVillage) and maize (CropDiseases).  
    (d) Classification accuracy across different synthetic-to-real ratios, showing consistent improvement with data augmentation.}
    \label{fig:imagegen_results}
\end{figure*}

Figure~\ref{fig:imagegen_results}(c) compares examples of real and generated images for each benchmark dataset. The synthetic tomato and maize leaves capture diverse disease phenotypes with realistic textures and lesion/pest patterns that closely resemble their real counterparts. Overall, the generated images are of high visual quality and in many cases are difficult to distinguish from real images.  
  
As shown in Figure~\ref{fig:imagegen_results}(d) and Table~\ref{tab:cnn-accuracies}, the classification model accuracy consistently improved with increasing synthetic ratios. For example, tomato classification accuracy rose from 0.779 to 0.843, and cassava accuracy from 0.602 to 0.765.

A general upward trend is observed, with accuracy improving as more synthetic images are incorporated into the training set. The models achieve their highest performance when the synthetic ratio reaches 400\% for all three species. Among the crops, tomato consistently exhibits lower accuracy compared to potato and bell pepper. This discrepancy arises because tomato classification is inherently more challenging, involving ten phenotypes, whereas potato and bell pepper classification involves only three and two phenotypes, respectively.

Overall, performance on CropDisease is lower than on PlantVillage.  This can be attributed to the greater difficulty of the CropDisease classification task, which requires distinguishing not only between viral and bacterial infections but also between pest-induced damage and abiotic stress symptoms. Another key observation is that the performance curves tend to plateau at higher synthetic ratios, suggesting that model capacity or noise in the dataset imposes an upper bound on achievable accuracy.  
Detailed classification results are provided in Table~\ref{tab:cnn-accuracies}.  

\begin{table}[t]
\centering
%\scriptsize
\setlength{\tabcolsep}{3.5pt}
\caption{Classification accuracies of the custom CNN model on two benchmark datasets under varying synthetic data ratios.}
\label{tab:cnn-accuracies}
\begin{tabular}{lcccccc}
\hline
\textbf{Dataset/Crop} & \textbf{0\%} & \textbf{50\%} & \textbf{100\%} & \textbf{200\%} & \textbf{300\%} & \textbf{400\%} \\
\hline
\multicolumn{7}{l}{\textbf{PlantV.}} \\
\hline
Bell pepper & 0.930 & 0.936 & 0.968 & 0.987 & 0.978 & 0.984 \\
Potato      & 0.919 & 0.950 & 0.957 & 0.938 & 0.950 & 0.973 \\
Tomato      & 0.779 & 0.814 & 0.821 & 0.822 & 0.838 & 0.843 \\
\hline
\multicolumn{7}{l}{\textbf{CropD.}} \\
\hline
Cassava     & 0.602 & 0.736 & 0.728 & 0.734 & 0.759 & 0.765 \\
Cashew      & 0.693 & 0.776 & 0.791 & 0.829 & 0.798 & 0.815 \\
Maize       & 0.546 & 0.574 & 0.692 & 0.667 & 0.682 & 0.700 \\
\hline
\end{tabular}
\end{table}

\subsection{Discussion}

Fine-tuning Stable Diffusion on agricultural imagery proved effective for generating high-fidelity plant images that capture phenotype-specific traits such as leaf discoloration, disease lesions, and pest damage. 
The resulting synthetic images were visually realistic and, in many cases, difficult to distinguish from real samples. 
The text-conditioned nature of the model further enabled flexible and controllable image synthesis, allowing targeted generation of crops under specified conditions (\textit{e.g.}, developmental stage or stress type). 
These properties make diffusion-based generation particularly well suited for agricultural applications, where data collection in outdoor is costly laborious.

Downstream classification experiments demonstrated that synthetic images can serve as effective data augmentation, consistently improving model performance across multiple benchmark datasets. 
However, performance gains saturated at higher synthetic-to-real ratios, suggesting diminishing returns once model capacity or dataset noise becomes the limiting factor. 
This highlights the importance of balancing real and synthetic data rather than maximizing synthetic volume indiscriminately.

We also observed domain-specific biases, particularly for outdoor imagery, which exhibited higher FID scores than indoor images. 
This reflects the greater variability and complexity of field environments and underscores the need for careful dataset curation and prompt design when adapting generative models to outdoor agricultural domains. 

Overall, the findings in this section emphasize that while general-purpose diffusion models provide a strong foundation, domain-specific fine-tuning and high-quality captioning are essential to capture the subtle visual characteristics required for reliable agricultural image synthesis.

%%%%%%%%%%%%%
\section{Text-Conditioned Image Translation}
The image translation component builds upon the domain-adapted diffusion model developed for text-conditioned image generation. 
Indoor plant images are abundant and easy to acquire under controlled conditions, whereas outdoor field imagery is much harder to collect and annotate, yet it is essential for deploying models in real-world agricultural settings. Bridging this indoor–to-outdoor gap via image translation can therefore expand field training data without costly field collections. Leveraging the diffusion model’s learned association between plant structures and textual semantics, our goal is to translate indoor plant images into realistic outdoor field scenes. 

\subsection{Datasets}
The indoor plant dataset used for image translation is derived from the study by Beck \textit{et al.}~\citep{beck2020embedded}, which provides high-resolution laboratory images of multiple crop species. 
For this task, we focus on canola as the target crop. 
Because the identifier token \texttt{sks} must encode only the plant’s structural semantics, indoor images were preprocessed to remove background elements such as pots and soil. 
Given the controlled imaging conditions and uniform blue background, plant segmentation was performed using a simple color-thresholding approach implemented with PlantCV~\citep{gehan2017plantcv}, an open-source toolkit widely used in plant phenotyping.

DreamBooth fine-tuning typically requires a small number of representative images (3–5) of the target subject captured from different viewpoints. 
Accordingly, segmented plant images were cropped and augmented using spatial transformations (\textit{e.g.}, flipping) to generate five inputs. 
These images were then used to fine-tune the diffusion model, with the rare token \texttt{sks} serving as the textual identifier for the target plant during training.

\subsection{Image Translation Method} %{DreamBooth-Based Text-Conditioned Image Translation}
To achieve structure-preserving translation, we adopt a text-inversion strategy within a DreamBooth \citep{ruiz2023dreambooth} fine-tuning framework. As shown in Figure~\ref{fig:overall_workflow}(a; third row), a rarely used identifier token (\textit{e.g.}, \texttt{sks})—chosen to avoid interference with existing semantic concepts in the pretrained vocabulary—is introduced to anchor the visual identity of the target plant. During training, this identifier token is injected into the text vocabulary and paired with images of the target plant, allowing the cross-attention layers of the latent diffusion model to associate \texttt{sks} with the plant’s visual structure.

Formally, DreamBooth adapts the standard denoising objective by conditioning the U-Net on text prompts augmented with the identifier token. The optimization objective can be written as
\begin{equation}
\mathcal{L}_{\text{DB}}(\theta) = \mathbb{E}_{z, \epsilon, t} 
\big\| \epsilon - \epsilon_\theta\big(z_t, t, \tau_\theta(y \cup w_{\text{sks}})\big) \big\|_2^2,
\end{equation}
where $y$ is the text prompt, $w_{\text{sks}}$ denotes the embedding of the identifier token, and $\tau_\theta(\cdot)$ is the text encoder. $z$ and $t$ are noised latent representation and time step respectively, $\epsilon$ and $\epsilon_{\theta}$ are true and predicted noises. To prevent catastrophic forgetting of general visual concepts, DreamBooth interleaves subject-specific fine-tuning with class prior preservation, ensuring that the model retains knowledge of the broader ``canola plant'' class, as shown in Section 2.3.

At inference time, the learned identifier token is combined with environment-specific descriptors to perform image translation. 
Prompts such as ``a sks canola plant growing in the TerraByte field'' enable the model to reconstruct the indoor plant structure while rendering an outdoor field environment consistent with the target domain. 

Unlike pixel-wise translation methods, this approach performs semantic conditioning within the diffusion process, yielding high-fidelity indoor-to-outdoor translations that preserve plant morphology while adapting environmental appearance in a controllable manner.

\subsection{Experiment Setup}
%\subsection{Model Training Setup}
DreamBooth fine-tuning for image translation was performed on three NVIDIA A100 GPUs using the captioned indoor canola dataset. 
Each target plant image was segmented, cropped, and augmented into five inputs for training. The model was fine-tuned for 400 steps using standard DreamBooth hyperparameters~\cite{ruiz2023dreambooth}, with the VAE and text encoder frozen while updating the U-Net backbone and the subject-specific token embedding associated with \texttt{sks}. 
At inference, translation outputs were controlled through text prompts that specify environmental and phenotypic attributes such as soil appearance, pest presence, plant arrangement, and developmental stage. 

\subsection{Evaluation Metrics and Results}
 The image translation framework was evaluated by visual inspection, quantitative metrics and downstream machine learning tasks.

\subsubsection{Qualitative Translation Results}
At inference time, different text prompts are used to condition the translation process, as shown in Figure~\ref{fig:translation_results}(a).  
By combining the token \texttt{sks} with environmental descriptors such as “a photo of a single sks canola plant taken in the outdoor TerraByte field,” the model generates realistic outdoor canola images that preserve the plant’s morphology while adapting the background, lighting, and soil textures of our target outdoor domain.  
Varying prompts enable flexible control over environmental and biological factors as seen in Figure~\ref{fig:translation_results}(a). For example, prompts describing arrangement of the plants (\textit{e.g.}, “sks canola plants growing in rows.”) introduce row-arrangement for the target plants. Environmental cues (\textit{e.g.}, “dark, moist soil” vs. “dry, light soil”) alter lighting contrast and soil colors. Developmental stage prompts (\textit{e.g.}, “a flowering sks canola plant”) produce plants with yellow flowers and more mature features.

These examples demonstrate that the model successfully recombines plant structure and environmental context through text conditioning, producing realistic and diverse outdoor translations.

\subsubsection{Quantitative Evaluation}
To complement qualitative analysis, we quantitatively assessed the realism and diversity of the translated canola images using the FID and IS.
The translated image set achieved an average FID of 162.1 and an IS of 2.48, reflecting good visual quality and diversity despite the challenging indoor-to-outdoor domain shift.
While the FID value is higher than that of directly generated outdoor images (162.1 vs.\ 127.6), indicating greater translation complexity, the IS remains comparable (2.48 vs.\ 2.60), suggesting that the translated outputs preserve semantic coherence and exhibit meaningful variability.

\subsection{Downstream Machine Learning Evaluation}
To further validate our image translation approach, we designed a downstream machine learning experiment to assess whether translated images can improve the performance of an object detection and classification model through data augmentation. Specifically, we evaluated the impact of using translated indoor soybean images (converted into outdoor scenes) to complement real field images for crop and weed detection and classification. This approach focuses on weed detection and identification at the species level, not only the general category.

\subsubsection{Dataset and Experiment Setup}
We designed a weed detection and classification experiment using three data sources. 
First, a benchmark dataset of 10,371 outdoor soybean field images~\citep{beck2020embedded} was used as the primary reference set, capturing realistic variability in lighting, background, and occlusion. 
Second, approximately 68,000 indoor soybean images collected under controlled laboratory conditions~\citep{beck2020embedded} served as inputs for generating translated outdoor-like soybean scenes. 
Third, cropped images of three common weed species---green foxtail (\textit{Setaria viridis} L.), redroot pigweed (\textit{Amaranthus retroflexus} L.), and kochia (\textit{Bassia scoparia} L.)—--were used to construct labeled detection datasets by overlaying weeds onto both real and translated soybean images.

\begin{figure*}[h]
\centering
\includegraphics[width=\linewidth]{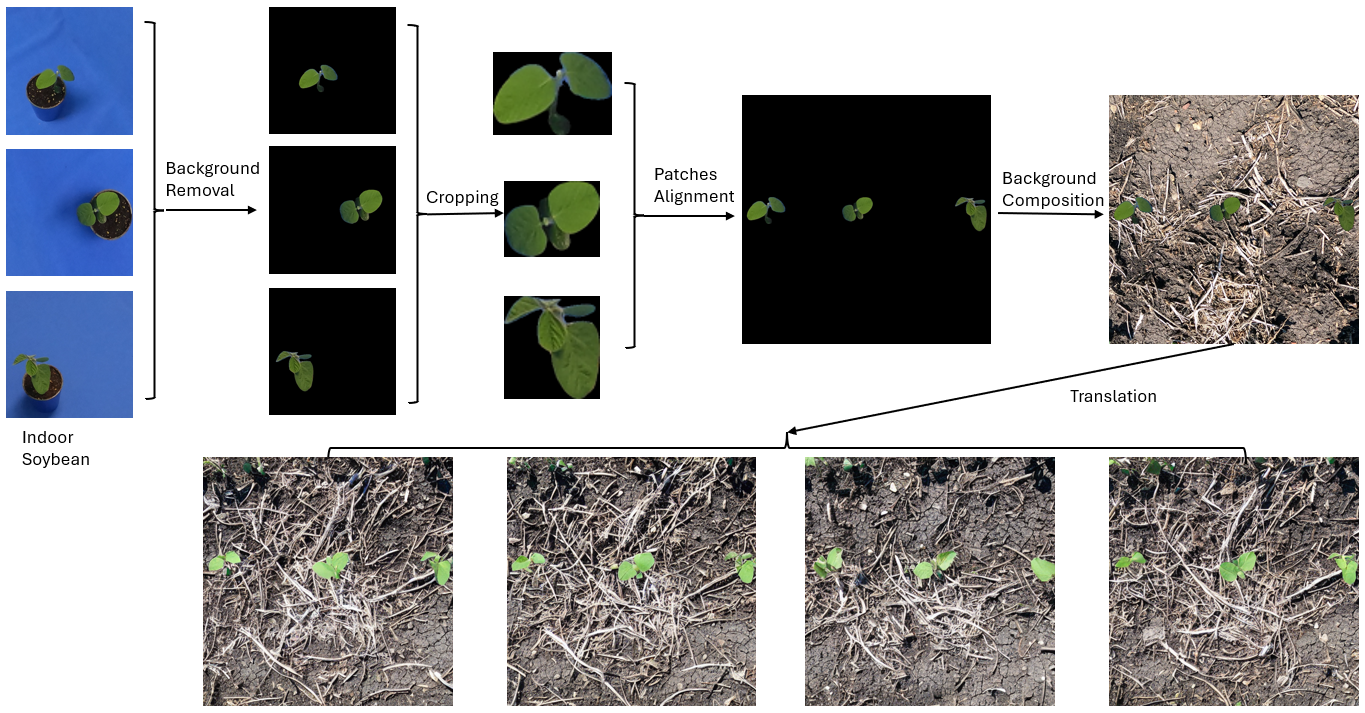}
\caption[ Translation workflow for the downstream image translation task.]{Overview of the dataset construction and translation workflow for the downstream indoor-to-outdoor soybean image translation. 
The process consists of: (1) removing background of the indoor soybean images, (2) cropping high quality plant patches, (3) compositing multiple patches onto a background canvas to mimic field-like arrangements, and 
(4) generating realistic outdoor translations using an image-conditioned diffusion model.}
\label{fig:translationDS}
\end{figure*}

Because text-conditioning alone is insufficient for preserving the spatial structure of multiple objects within a single image, we adopted an image-guided conditional generation strategy~\citep{meng2021sdedit} for this task. Indoor soybean images, which typically contain a single plant, were first segmented and cropped using bounding boxes obtained from YOLOv8~\citep{yolov8}, followed by background removal with PlantCV. After filtering noisy samples, approximately 40,000 high-quality soybean patches were retained. These patches were then composited onto empty canvases in row-like arrangements (3–6 plants per image), with outdoor background images inserted to mimic field conditions. The resulting composite images were translated into outdoor environments using the image-conditioned diffusion model. Weeds were subsequently overlaid onto both translated and real soybean images using identical procedures to ensure consistency. Weeds included three representative species: green foxtail (\textit{Setaria viridis} L.), redroot pigweed (\textit{Amaranthus retroflexus} L.), and kochia (\textit{Bassia scoparia} L.). Details of this data preparation are demonstrated in Figure~\ref{fig:translationDS}.

To assess the impact of translated data, mixed datasets were constructed by augmenting real outdoor soybean–weed images with translated counterparts at synthetic ratios of 0\%, 50\%, 100\%, 150\%, 200\%, and 300\%. For each ratio, a YOLOv8 detector (\citep{yolov8}) was fine-tuned for joint soybean and weed detection and classification. YOLOv8 was selected due to its fast training speed and strong baseline performance, making it suitable for evaluating performance trends under varying data augmentation levels rather than achieving state-of-the-art accuracy. 
All models were evaluated on a held-out test set containing only real soybean–weed images, using standard metrics including precision, recall, and mAP50, which is a standard metric for object detection measuring the overlapping between true and predicted bounding boxes.

\subsubsection{Image-Guided Image Translation}
While DreamBooth performs well for single-plant scenes, it struggles with multi-object compositions. Therefore, for downstream detection tasks, we used Stable Diffusion's image-to-image (img2img) pipeline~\citep{meng2021sdedit} for image-guided translation. This approach refines an existing composited image rather than generating from scratch. The process initializes the denoising trajectory from a partially noised version of the input image. In our implementation, single-plant indoor soybean patches were segmented, cropped, and composited onto real outdoor soil backgrounds to simulate realistic multi-plant scenes.  
The img2img pipeline then refined these composited images into realistic outdoor field scenes (Figure\ref{fig:translationDS}). This process produced translated images suitable for augmenting object detection and classification datasets involving both soybean and the three weed species.

\subsubsection{Results and Analysis} 
\begin{figure*}[t]
    \centering
    \includegraphics[width=\linewidth]{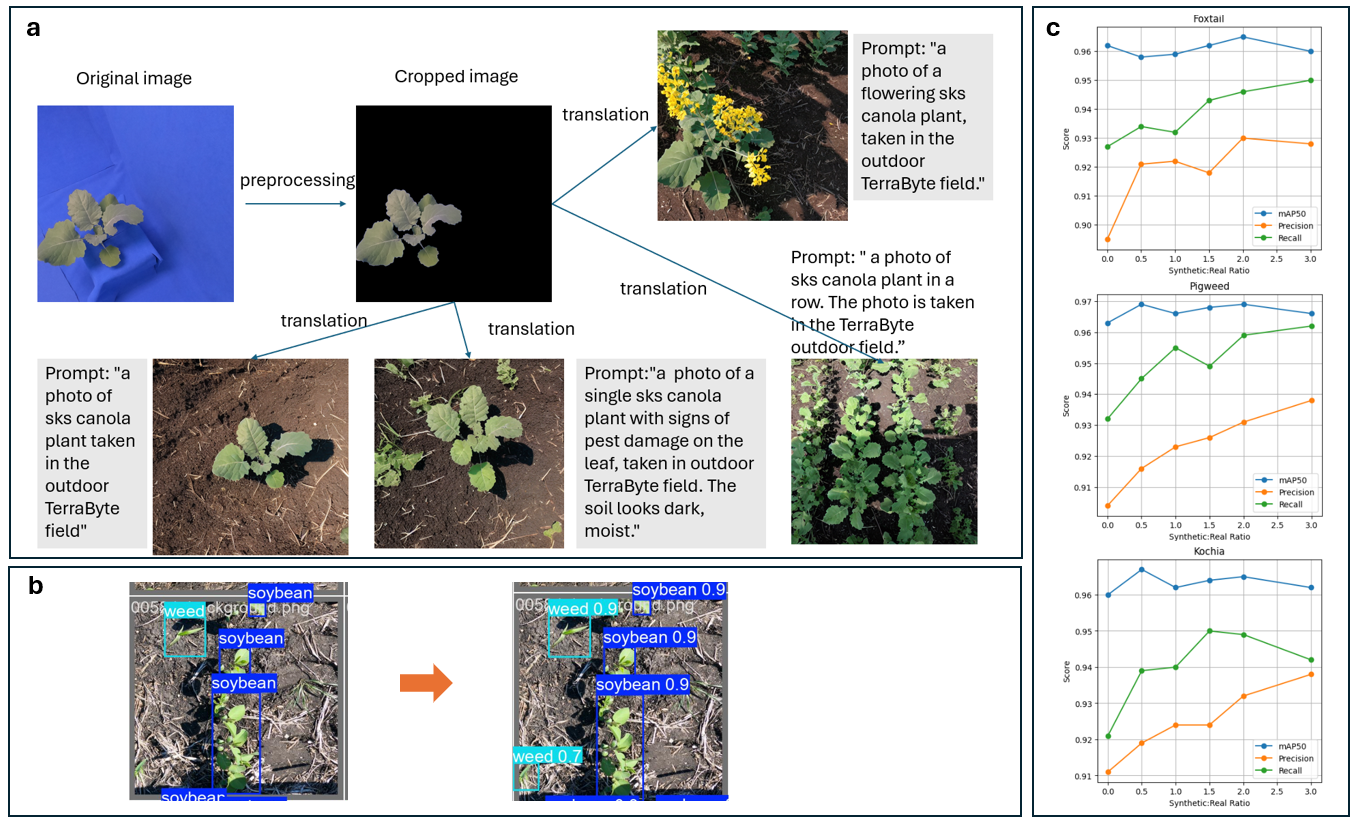}
    \caption{\textbf{Indoor-to-Outdoor Image Translation and Downstream Detection.}    
    (a) Examples of translated images under different text prompts simulating environmental variations (lighting, soil, developmental stage, arrangement).  
    (b) YOLOv8n detection and classification examples showing accurate bounding boxes for soybean and weed plants.  
    (c) Detection metrics (precision, recall, and mAP50) across three weed species and varying synthetic-to-real ratios.  
    Translation-based augmentation consistently improves model performance.}
    \label{fig:translation_results}
\end{figure*}

Figure~\ref{fig:translation_results}(b) presents detection examples showing accurate bounding boxes and class labels for both soybean and weed plants. Table~\ref{tab:yolo_performance} summarizes the mean precision, recall, and mAP50 scores for each synthetic ratio and weed species. Overall, we observe consistently high detection accuracy for all weed species, with mAP50 values averaging around 0.96. For precision and recall, performance improved consistently with higher synthetic ratios. For example, as shown in Figure~\ref{fig:translation_results}(c), precision for pigweed group increased from 0.904 at 0\% synthetic data to 0.931 at 200\%, to 0.938 at 300\% respectively. These results demonstrate that diffusion-based translation can effectively augment real datasets for field-vision tasks.

\begin{table}[h]
\centering
\caption{Performance of YOLOv8n on weed detection and classification under different synthetic ratios.}
\label{tab:yolo_performance}
\resizebox{\linewidth}{!}{
\begin{tabular}{lcccccc}
\hline
\textbf{Species} & \textbf{Metric} & \textbf{0\%} & \textbf{50\%} & \textbf{100\%} & \textbf{200\%} & \textbf{300\%} \\
\hline
\multirow{3}{*}{Foxtail} 
& Precision & 0.895 & 0.921 & 0.922 & 0.930 & 0.928 \\
& Recall    & 0.927 & 0.934 & 0.932 & 0.946 & 0.950 \\
& mAP50    & 0.962 & 0.958 & 0.959 & 0.965 & 0.960 \\
\hline
\multirow{3}{*}{Pigweed} 
& Precision & 0.904 & 0.916 & 0.923 & 0.931 & 0.938 \\
& Recall    & 0.932 & 0.945 & 0.955 & 0.959 & 0.962 \\
& mAP50    & 0.963 & 0.969 & 0.966 & 0.969 & 0.966 \\
\hline
\multirow{3}{*}{Kochia} 
& Precision & 0.911 & 0.919 & 0.924 & 0.932 & 0.938 \\
& Recall    & 0.921 & 0.939 & 0.940 & 0.949 & 0.942 \\
& mAP50    & 0.960 & 0.967 & 0.962 & 0.965 & 0.962 \\
\hline
\end{tabular}}
\end{table}

\subsection{Discussion}
The image translation section reveals complementary strengths of text-conditioned and image-guided diffusion models for agricultural image translation. DreamBooth enables precise, prompt-driven control by binding rare tokens (e.g., \texttt{sks}) to plant structures, making it effective for translating single-plant images while preserving morphology. However, it struggles with multi-object field scenes, often failing to maintain spatial consistency when multiple plants are present.

The image-guided img2img approach proved more robust for agricultural scenes with multiple plants in our study. 
By conditioning generation on composited indoor images, it preserved plant arrangements while adapting environmental appearance. Its main limitation lies in its dependence on preprocessing quality, as segmentation and compositing artifacts can affect translation realism.

Downstream experiments demonstrate that translated images improve detection performance when combined with real data, particularly in recall and precision. However, performance gains saturate at high synthetic ratios, indicating that balanced augmentation is more important than data volume. Overall, the results suggest that translation strategy should be chosen based on scene complexity.

%%%%%%%%%%%%%%%
\section{Preference-Guided Fine-Tuning}
The final component of our framework introduces expert preference into the diffusion training process.While standard diffusion models can generate visually plausible images, their outputs are not necessarily aligned with botanical realism and structural quality. To address this gap, we developed a two-stage preference-guided fine-tuning pipeline that incorporates a learned reward model to bias the diffusion process toward expert-preferred outputs.

\subsection{Materials and Method}
As shown in Figure~\ref{fig:reward_finetune_results}(a), to align the generative model with expert judgment (even though we used the term ``expert'' here, we scored these images as a proof of concept), we implemented a reward-based fine-tuning pipeline. A convolutional reward model was trained on 500 manually scored images generated by our domain-adapted Stable Diffusion model from Section 4, where each score $r \in [0,10]$ reflected perceived realism. The reward model operated in latent space ($4 \times 64 \times 64$) extracted from the Sable Diffusion VAE encoder and achieved a Pearson correlation of 0.79 with expert scoring. The reward model was optimized using mean-squared error loss between predicted and annotated scores:
\begin{equation}
    \mathcal{L}_{\text{reward}} = \frac{1}{N} \sum_{i=1}^N (\hat{r}_i - r_i)^2.
\end{equation}

The reward model is implemented as a CNN that progressively reduces the spatial dimensions of the latent tensor while increasing feature channels. Feature aggregation is achieved through convolutional blocks with batch normalization and ReLU activations, followed by global average pooling. The final fully connected head maps the aggregated features to a scalar value representing the predicted reward. The full architecture is summarized in Appendix Table~\ref{tab:reward_model_arch}. 

During fine-tuning, the model generated $N=12$ candidate images per prompt. The reward model predicted scores $\{r_1, \ldots, r_N\}$, and the top-$k$ ($k=8$) samples were selected. A reward-weighted denoising loss was computed:
\begin{equation}
    \mathcal{L}_{\text{SFT}} = \sum_{i=1}^k w_i \| \epsilon_\theta(z_t^i, t) - \epsilon \|^2,
\end{equation}
where 
\begin{equation}
    w_i = \exp(r_i / \tau) / \sum_j \exp(r_j / \tau),
\end{equation}
 is normalized preference weight, $\tau$ is a temperature hyperparameter controlling the sharpness of the weighting distribution, $\epsilon_\theta$ is the U-Net’s predicted noise, $\epsilon$ is the true Gaussian noise, and $z_t^i$ is the noisy latent sampled at timestep $t$ using the diffusion forward process. %The model was optimized using AdamW with a learning rate of $1\times10^{-5}$ and EMA weight decay of $0.999$.
The U-Net denoiser was fine-tuned using the AdamW optimizer~\citep{loshchilov2017decoupled} with a learning rate of $1 \times 10^{-5}$, and gradient clipping set to $0.5$. An exponential moving average of model weights was maintained with a decay factor of $0.999$ to improve training stability. We used $\tau = 1$ in our training process. Validation was performed at the end of each epoch by generating $M=50$ images from a held-out set of prompts. These images were scored by the reward model, and the mean reward score was recorded. The best model checkpoint was selected based on the highest validation mean reward.

\subsection{Evaluation and Results}
\begin{figure*}[t]
    \centering
    \includegraphics[width=\linewidth]{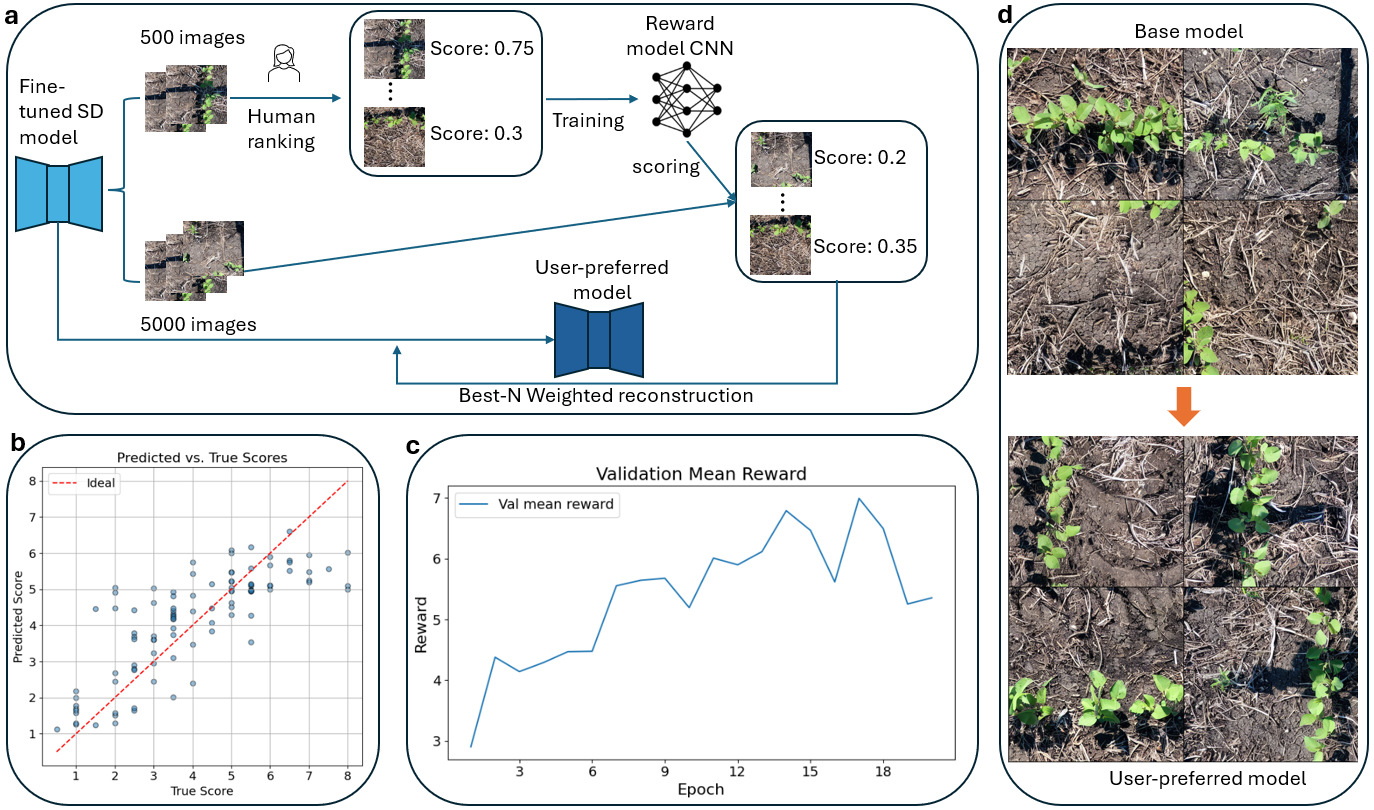}
    \caption{\textbf{Preference-Guided Fine-Tuning of the Diffusion Model.}  
    (a) Workflow of the two-stage reward alignment pipeline integrating expert scoring, reward modeling, and weighted supervised fine-tuning.  
    (b) Quantitative results showing the scatter of predicted vs.\ true rewards.
    (c) The training curve of mean reward score across epochs in preference-guide model fine-tuning.  
    (d) Visual comparison between base and preference-guided models for identical prompts.}
    \label{fig:reward_finetune_results}
\end{figure*}

Figure~\ref{fig:reward_finetune_results}(b) shows the scatter plot of predicted versus true reward scores for the test set, demonstrating a strong positive correlation (Pearson $r = 0.79$). This confirms that the reward model effectively captured expert preference patterns in the image quality scores.  
Figure~\ref{fig:reward_finetune_results}(c) illustrates the evolution of the mean reward during fine-tuning, which consistently increased and maxed around epoch 17, indicating successful integration of the reward signal and progressive improvement in generated image quality.  
Figure~\ref{fig:reward_finetune_results}(d) qualitatively compares images produced by the base (before tuning) and preference-aligned (tuned) models using identical prompts.  
The tuned model generated outputs with improved structural consistency while reducing artifacts such as leaf deformation or off-centered plants commonly seen in the base model.  
Although the tuned model exhibited a higher FID score (189.3 vs.\ 108.7 when compared to real outdoor images), the mean reward score increased substantially (6.88 vs.\ 3.72), and a paired $t$-test confirmed that this difference was statistically significant ($t=10.98$, $p = 1.42 \times 10^{-17}$). This confirms that preference-guided optimization enhanced subjective realism and overall stability despite slight divergence from the training data distribution.

\subsection{Discussion}
This section demonstrated that preference-guided fine-tuning can effectively align diffusion model outputs with expert-defined quality criteria by integrating a learned reward model into the training process. Using a reward-weighted supervised fine-tuning strategy, the preference-aligned model achieved substantially higher reward scores than the base model, confirming that subjective notions of image quality can be successfully encoded and optimized. 

However, this improvement came with a trade-off: the tuned model exhibited higher FID relative to real images, indicating a shift in the generative distribution away from statistical fidelity. 
Such divergence may be partially beneficial, as it reduces the prevalence of low-quality outputs and concentrates generation toward expert-preferred samples. At the same time, the approach inherits limitations from the reward model itself, including sensitivity to annotation quality and potential bias, as reflected by the imperfect correlation between predicted and true scores. 

Despite these challenges, preference-guided fine-tuning offers a practical and flexible mechanism for incorporating expert feedback into generative modeling, particularly in domains such as agriculture where image quality is difficult to define objectively.

\section{Overall Discussion and Conclusion}

This paper investigated the use of diffusion-based generative models to address data scarcity in agricultural computer vision.  The following discussion synthesizes the key findings, practical implications, limitations, and future directions of this work.

\subsection{Summary of Contributions}
This work contributes to agricultural AI in three complementary directions. First, we demonstrated that text-conditioned fine-tuning of Stable Diffusion models enables realistic generation of crop imagery across developmental stages and conditions, with synthetic data improving downstream phenotype classification performance.  
Second, we explored indoor-to-outdoor image translation using both DreamBooth-based text conditioning and image-guided diffusion, showing that translated images can effectively augment training data for weed detection and classification. Third, we introduced a preference-aligned fine-tuning strategy using a learned reward model, enabling diffusion models to better align outputs with expert-defined notions of quality and realism.  

Together, these contributions illustrate how generative modeling can be systematically adapted, evaluated, and aligned for agricultural applications.

\subsection{Effectiveness of Diffusion Models for Agricultural Imagery}
Across all tasks, diffusion models proved effective at modeling complex agricultural imagery. Fine-tuned text-to-image models produced realistic plant structures and phenotypes, while translated images successfully bridged the gap between controlled indoor data and variable outdoor field scenes. Importantly, synthetic and translated images were not merely visually plausible but demonstrably useful: mixing them with real data consistently improved downstream performance in both classification and detection tasks.  

These results indicate that diffusion models can serve as practical data augmentation tools in agriculture, particularly when real-world data collection is limited by cost, seasonality, or annotation effort.

\subsection{Trade-offs Between Objective Metrics and Subjective Quality}
A noteworthy observation was the divergence between objective generative metrics and human judgment. While metrics such as FID and IS provide useful quantitative benchmarks, they did not always align with expert perception or downstream utility. Preference-aligned fine-tuning, for example, improved expert-rated image quality and stability but resulted in worse FID scores, reflecting a shift in the generative distribution.

This highlights the need for multi-dimensional evaluation in agricultural AI. Effective assessment should combine objective similarity metrics, expert preference alignment, and task-level performance to ensure that generated data is both statistically meaningful and practically useful.

\subsection{Practical Implications for Agricultural AI}
The findings of this work have several practical implications.  
Generative diffusion models can substantially reduce reliance on costly field data collection by augmenting scarce datasets with realistic synthetic imagery. This capability accelerates experimentation and deployment of machine learning models in agriculture.  

Moreover, the ability to align generative outputs with expert preferences introduces a user-centered dimension to generative pipelines. Such adaptability is particularly valuable in agricultural contexts, where realism is often defined by domain expertise rather than strict statistical similarity. Preference-guided generation may therefore support applications such as digital phenotyping and decision support systems for agriculture.

\subsection{Limitations and Challenges}
Despite promising results, several limitations remain.  
Generated and translated images exhibited variability in quality, including occasional artifacts related to lighting, shadows, or structural consistency. Preference-aligned fine-tuning introduced trade-offs between subjective quality and distributional fidelity, underscoring challenges in balancing realism and alignment. In addition, the reward model relied on manual expert annotation, limiting scalability and introducing potential bias. Finally, the methods were primarily validated on canola and soybean imagery, and broader evaluation across diverse crops and environments is required to assess generalizability.

\subsection{Future Directions}
Future work can extend this framework in several ways.  
First, larger vision--language models may improve caption quality and conditioning robustness. Second, hybrid translation strategies that combine DreamBooth’s semantic control with image-guided structural preservation could better handle complex field scenes. Finally, integrating generative models with structured agronomic knowledge could further constrain outputs to biologically and environmentally plausible settings, increasing reliability for real-world agricultural applications. 

\subsection{Conclusion}
This work demonstrates that diffusion-based generative models provide an effective and flexible solution to data scarcity in agricultural vision tasks. By unifying text-conditioned image generation, indoor-to-outdoor translation, and preference-aligned fine-tuning within a single framework, the proposed pipeline enables the creation of realistic, task-relevant synthetic imagery that improves downstream model performance and robustness. The results show that synthetic data generated by fine-tuned diffusion models can meaningfully complement real-world datasets, particularly when domain gaps and annotation limitations are present.

Beyond data augmentation, this study highlights the importance of incorporating expert feedback into generative modeling. Preference-aligned fine-tuning steers diffusion models toward outputs that better reflect domain-specific quality criteria, improving stability and subjective realism. Together, these findings position diffusion-based generative pipelines as a foundation for data-efficient, expert-aware agricultural AI systems, with implications for plant phenotyping and precision agriculture in real-world farming environments.

%%%%%%%%%%%%%%
\section*{CRediT Authorship Contribution Statement}
\textbf{Da Tan:} Conceptualization, Methodology, Software, Validation, Formal analysis, Investigation, Visualization, Writing - Original draft. \textbf{Michael Beck:} Data Curation, Writing - Review \& Editing. \textbf{Christopher Bidinosti:} Conceptualization, Data Curation, Writing - Review \& Editing, Project administration. \textbf{Robert Gulden:} Formal analysis, Writing - Review \& Editing. \textbf{Christopher Henry:} Conceptualization, Methodology, Validation, Formal analysis, Resources, Writing - Review \& Editing, Supervision, Project Administration, Funding acquisition.

\section*{Declaration of Competing Interest}
The authors declare no competing financial interests or personal relationships that could have influenced the work reported in this paper.

\section*{Acknowledgments}
This research was supported by the University of Manitoba Graduate Fellowship (UMGF), the University of Manitoba Faculty of Science SEGS program, and NSERC Discovery Grant (RGPIN-2025-06057).

\section*{Data Availability}
The canola and soybean data in Section 3 and 4 is available using the \href{https://terrabyte.acs.uwinnipeg.ca/resources.html}{TerraByte Data Client}. 
The benchmark PlantVillage dataset is available at \href{https://www.kaggle.com/datasets/emmarex/plantdisease}{Kaggle}.
The benchmark CropDisease dataset is available at \href{https://data.mendeley.com/datasets/bwh3zbpkpv/1}{Mendeley}.
The code implementation will be publicly available.

\bibliographystyle{cas-model2-names}
\bibliography{cas-refs}

\FloatBarrier
% \onecolumn
% \appendix
% \appendixpage
% \section{APPENDIX}
% %\addcontentsline{toc}{section}{APPENDIX} % optional
%%%%%%%%%%%%
%\clearpage        % flush all pending floats
\onecolumn
\appendix
%[\section*{APPENDIX}]
\section*{APPENDIX}

\clearpage
\setcounter{figure}{0}
\setcounter{table}{0}

\renewcommand{\thefigure}{A\arabic{figure}}
\renewcommand{\thetable}{A\arabic{table}}

%%%%%%%%%%%%%%
%\onecolumn
\begin{figure}[H]%[t]
\centering
\includegraphics[width=\linewidth]{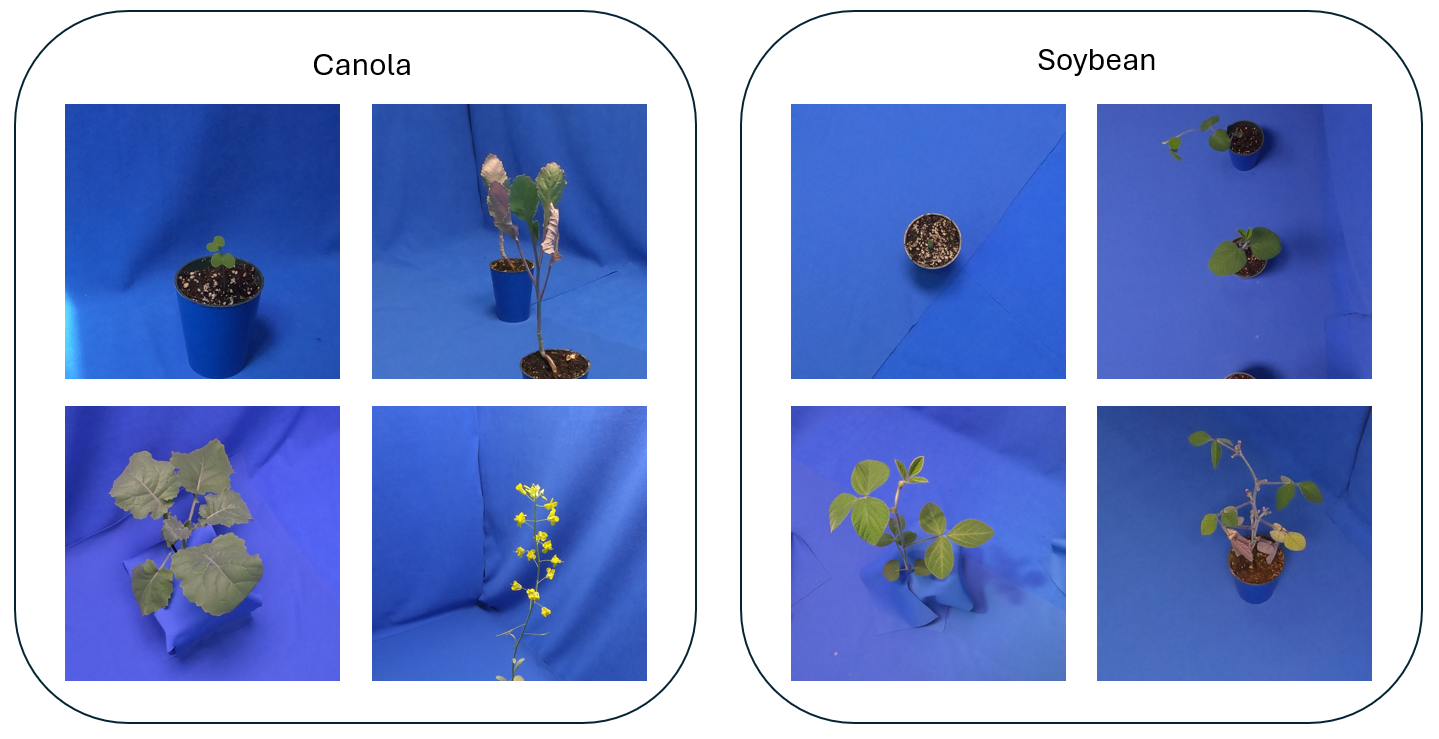}
\caption[Examples of indoor canola and soybean images.]{Examples of indoor canola and soybean images in different developmental stages or conditions.}
\label{fig:indoor_examples_canola}
\end{figure}

\begin{figure}[H]
\centering
\includegraphics[width=\linewidth]{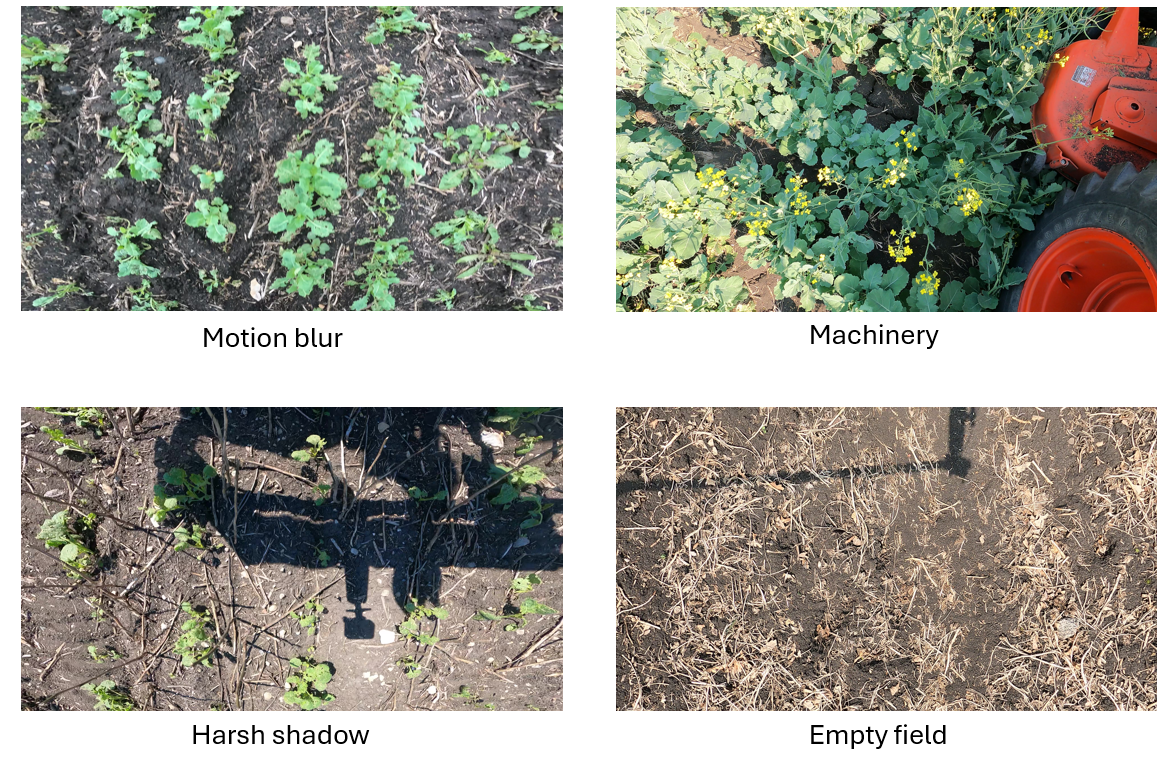}
\caption[Examples of low-quality outdoor canola images.]{Examples of low-quality outdoor canola images across different developmental stages and conditions, illustrating representative challenges such as motion blur, background machinery, harsh shadows, and empty fields.}
\label{fig:low_qua_outdoor_examples}
\end{figure}

\begin{figure}[H]
\centering
\includegraphics[width=\linewidth]{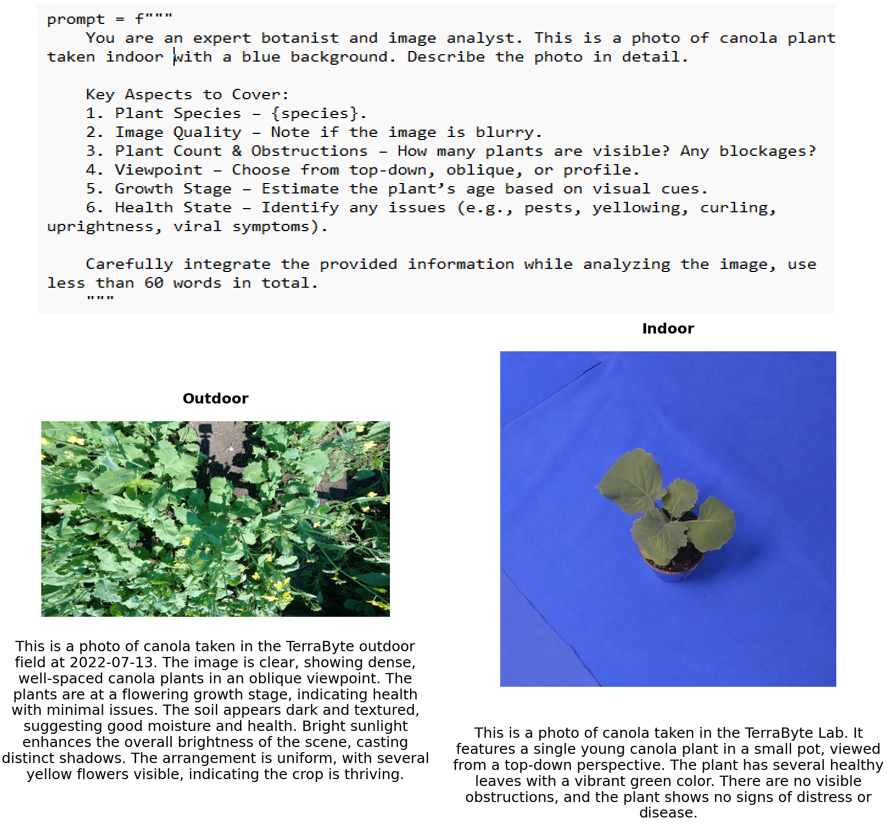}
\caption[Example of the prompt template and caption]{Example of the prompt template and caption generation process using ChatGPT-4o. 
Shown are sample indoor and outdoor canola plant images alongside their automatically generated captions. 
The prompt was designed to highlight plant species, developmental stage, and environmental context, ensuring informative conditioning text for fine-tuning Stable Diffusion model.}
\label{fig:prompt}
\end{figure}

%\captionof{table}%{H}
\begin{table}[H]
%\centering
\caption[Number of images for the PlantVillage dataset]{Number of images in the training and test sets for tomato, potato, and bell pepper phenotypes in the PlantVillage dataset.}
\label{tab:tomato-potato-bellpepper}
\begin{tabular}{lrrr}
\hline
\textbf{Tomato phenotypes} & \textbf{Train} & \textbf{Test} & \textbf{Total} \\
\hline
Healthy & 1012 & 434 & 1446 \\
Tomato mosaic virus & 259 & 111 & 370 \\
Leaf mold & 630 & 270 & 900 \\
Early blight & 689 & 296 & 985 \\
Target spot & 1031 & 442 & 1473 \\
Spotted spider mite & 1180 & 506 & 1686 \\
Septoria leaf spot & 1200 & 515 & 1715 \\
Late blight & 1377 & 590 & 1967 \\
Bacterial spot & 1512 & 648 & 2160 \\
Yellow leaf curl virus & 2230 & 957 & 3187 \\
\hline
\textbf{Potato phenotypes} & & & \\
\hline
Healthy & 117 & 50 & 167 \\
Early blight & 648 & 277 & 925 \\
Late blight & 739 & 317 & 1056 \\
\hline
\textbf{Bell pepper phenotypes} & & & \\
\hline
Healthy & 1056 & 453 & 1509 \\
Bacterial spot & 761 & 326 & 1087 \\
\hline
\end{tabular}
\end{table}
%\caption{table}

\begin{table}[H]
\centering
\caption[Number of images for the CropDiseases dataset]{Number of images in the training and test sets for maize, cashew, and cassava phenotypes in the CropDisease dataset.}
\label{tab:maize-cashew-cassava}
\begin{tabular}{lrrr}
\hline
\textbf{Maize phenotypes} & \textbf{Train} & \textbf{Test} & \textbf{Total} \\
\hline
Healthy & 121 & 52 & 173 \\
Fall armyworm & 196 & 84 & 280 \\
Grasshopper & 484 & 207 & 691 \\
Leaf beetle & 609 & 261 & 870 \\
Streak virus & 645 & 277 & 922 \\
Leaf blight & 711 & 305 & 1016 \\
Leaf spot & 860 & 369 & 1229 \\
\hline
\textbf{Cashew phenotypes} & & & \\
\hline
Healthy & 988 & 424 & 1412 \\
Gummosis & 280 & 120 & 400 \\
Leaf miner & 968 & 415 & 1383 \\
Red rust & 1194 & 513 & 1707 \\
Anthracnose & 1226 & 526 & 1752 \\
\hline
\textbf{Cassava phenotypes} & & & \\
\hline
Healthy & 760 & 326 & 1086 \\
Green mite & 875 & 375 & 1250 \\
Mosaic & 885 & 380 & 1265 \\
Brown spot & 1032 & 442 & 1474 \\
Bacterial blight & 1697 & 728 & 2425 \\
\hline
\end{tabular}
\end{table}

\begin{table}[H]
\centering
\caption[CNN architecture for the image generation machine learning tasks.]{Summary of CNN architecture used as the downstream machine learning task to assess synthetic image quality.}
\label{tab:cnn-architecture-compact}
\begin{tabular}{ll}
\hline
\textbf{Stage} & \textbf{Configuration} \\
\hline
Input & $3 \times 224 \times 224$ RGB image \\
Feature extractor & 
\begin{tabular}[t]{@{}l@{}}
Conv(32, $3 \times 3$) → BN → ReLU → MaxPool(2) \\
Conv(64, $3 \times 3$) → BN → ReLU → MaxPool(2) \\
Conv(128, $3 \times 3$) → BN → ReLU → MaxPool(2) \\
Conv(256, $3 \times 3$) → BN → ReLU → AdaptiveAvgPool(1)
\end{tabular} \\
Classifier & Flatten → Dropout(0.5) → Linear($256 \rightarrow$ num\_classes) \\
\hline
\end{tabular}
\end{table}

\begin{table}[H]
\centering
\caption[Architecture of the CNN-based reward model]{Architecture of the CNN-based reward model. The input is a latent tensor of shape $4 \times 64 \times 64$.}
\label{tab:reward_model_arch}
\begin{tabular}{llp{6cm}}
\hline
\textbf{Layer} & \textbf{Output Shape} & \textbf{Details} \\
\hline
Input & $4 \times 64 \times 64$ & Scaled VAE latent \\
Conv2d + BN + ReLU & $32 \times 32 \times 32$ & $4 \to 32$, kernel=3, stride=1, pool=2 \\
Conv2d + BN + ReLU & $64 \times 16 \times 16$ & $32 \to 64$, kernel=3, stride=1, pool=2 \\
Conv2d + BN + ReLU & $128 \times 8 \times 8$ & $64 \to 128$, kernel=3, stride=1, pool=2 \\
Conv2d + BN + ReLU & $256 \times 1 \times 1$ & $128 \to 256$, kernel=3, stride=1, global avg pool \\
Flatten & $256$ & -- \\
Linear + ReLU + Dropout & $128$ & Dropout $p=0.3$ \\
Linear & $1$ & Reward prediction \\
\hline
\end{tabular}
\end{table}

\end{document}